\documentclass[11pt,a4paper]{article}
\usepackage{amsmath}
\usepackage{graphicx}
\usepackage[hyperref]{emnlp2020}
\usepackage{times}
\usepackage{latexsym}
\usepackage{makecell}

\usepackage{enumerate}
\usepackage{enumitem}
\usepackage{soul}

\usepackage{xurl}
\urldef{\kaggleurl}\url{https://www.kaggle.com/c/jigsaw-unintended-bias-in-toxicity-classification}

\usepackage{microtype}

\aclfinalcopy 

\usepackage{multirow}
\DeclareMathOperator*{\argmin}{\arg\min}
\DeclareMathOperator*{\argmax}{\arg\max}

\newcommand{\E}{{\rm I\kern-.3em E}}

\newcommand{\citenoun}[1]{{\citet{#1}}}
\newcommand{\cut}[1]{}

\newenvironment{packed_enum}{
\begin{enumerate}[topsep=.07cm,leftmargin=.2in]
  \setlength{\itemsep}{1pt}
  \setlength{\parskip}{0pt}
  \setlength{\parsep}{0pt}
}{\end{enumerate}}

\newenvironment{packed_item}{
\begin{itemize}[topsep=.07cm,leftmargin=.2in]
   \setlength{\itemsep}{1pt}
   \setlength{\parskip}{0pt}
   \setlength{\parsep}{0pt}
}{\end{itemize}}

\title{Identifying Spurious Correlations for Robust Text Classification}

\author{Zhao Wang \\
      Illinois Institute of Technology \\
      \texttt{zwang185@hawk.iit.edu} \\\And
      Aron Culotta \\
      Tulane University \\
      \texttt{aculotta@tulane.edu} \\}

\date{}

\begin{document}
\maketitle
\begin{abstract}
The predictions of text classifiers are often driven by spurious correlations -- e.g., the term {\it Spielberg} correlates with positively reviewed movies, even though the term itself does not semantically convey a positive sentiment. In this paper, we propose a method to distinguish spurious and genuine correlations in text classification. We treat this as a supervised classification problem, using features derived from treatment effect estimators to distinguish spurious correlations from ``genuine'' ones. Due to the generic nature of these features and their small dimensionality, we find that the approach works well even with limited training examples, and that it is possible to transport the word classifier to new domains. Experiments on four datasets (sentiment classification and toxicity detection) suggest that using this approach to inform feature selection also leads to more robust classification, as measured by improved worst-case accuracy on the samples affected by spurious correlations.
\end{abstract}

\section{Introduction}
\label{sec.intro}
Text classifiers often rely on spurious correlations. For example, consider sentiment classification of movie reviews. The term {\it Spielberg} may be correlated with the positive class because many of director Steven Spielberg's movies have positive reviews. However, the term itself does not indicate a positive review. In other words, the term {\it Spielberg} does not {\it cause} the review to be positive. Similarly, consider the problem of toxicity classification of online comments. Terms indicative of certain ethnic groups may be associated with the toxic class because those groups are often victims of harassment, not because those terms are toxic themselves.

Oftentimes, such spurious correlations do not harm prediction accuracy because the same correlations exist in both training and testing data (under the common assumption of i.i.d. sampling). However, they can still be problematic for several reasons. For example, under {\it dataset shift}~\cite{quionero2009dataset}, the testing distribution differs from the training distribution. E.g., if Steven Spielberg makes a new, bad movie, the sentiment classifier may incorrectly classify the reviews as positive because they contain the term {\it Spielberg}. Additionally, if the spurious correlations indicate demographic attributes, then the classifier may suffer from issues of {\it algorithmic fairness}~\cite{kleinberg2018algorithmic}. For example, the toxicity classifier may unfairly over-predict the toxic class for comments discussing certain demographic groups. Finally, in settings where classifiers must explain their decisions to humans, such spurious correlations can reduce trust in autonomous systems~\cite{guidotti2018survey}.

In this paper, we propose a method to distinguish {\it spurious} correlations, like {\it Spielberg}, from {\it genuine} correlations, like {\it wonderful}, which more reliably indicate the class label. Our approach is to treat this as a separate classification task, using features drawn from treatment effect estimation approaches that isolate the impact each word has on the class label, while controlling for the context in which it appears. 

We conduct classification experiments with four datasets and two tasks (sentiment classification and toxicity detection), focusing on the problem of {\it short text} classification (i.e., single sentences or tweets). We find that with a small number of labeled word examples (200-300), we can fit a classifier to distinguish spurious and genuine correlations with moderate to high accuracy (.66-.82 area under the ROC curve), even when tested on terms most strongly correlated with the class label.  In addition, due to the generic nature of the features, we can train a word classifier on one domain and transfer it to another domain without much loss in accuracy. 

Finally, we apply the word classifier to inform feature selection for the original classification task (e.g., sentiment classification and toxicity detection). Following recent work on distributionally robust classification~\cite{sagawa2020distributionally}, we measure worst-case accuracy by considering samples of data most affected by spurious correlations. We find that removing terms in the order of their predicted probability of being spurious correlations can result in more robust classification with respect to this worst-case accuracy.



\section{Problem and Motivation}
\label{sec.problem}

We consider binary classification of short documents, e.g., sentences or tweets. Each sentence is a sequence of words $s=\langle w_1 \ldots w_k \rangle$ with a corresponding binary label $y \in \{-1, 1\}$. To classify a sentence $s$, it is first transformed into a feature vector $x$ via a feature function $g: s \mapsto x$. Then, the feature vector is assigned a label by a classification function $f : (x; \theta) \mapsto \{-1,1\}$, with model parameters $\theta$. Parameters $\theta$ are typically estimated from a set of i.i.d. labeled examples $\mathcal{D}=\{(s_1,y_1) \ldots (s_n,y_n)\}$ by minimizing some loss function $\mathcal{L}$:  $\theta^* \leftarrow \argmin_{\theta} \mathcal{L}(\cal{D}, \theta)$.

To illustrate the problem addressed in this paper, we will first consider the simple approach of a bag-of-words logistic regression classifier. In this setting, the feature function $g(s)$ simply maps a document to a word count vector $x=\{x_1 \ldots x_V\}$, for vocabulary size $V$, and the classification function is the logistic function $f(x; \theta) = \frac{1}{1+e^{-\langle x, \theta \rangle}}$. After estimating parameters $\theta$ on labeled data $\mathcal{D}$, we can then examine the coefficients corresponding to each word in the vocabulary to see which words are most important in the model.

\begin{figure}[t]
\includegraphics[width=\linewidth]{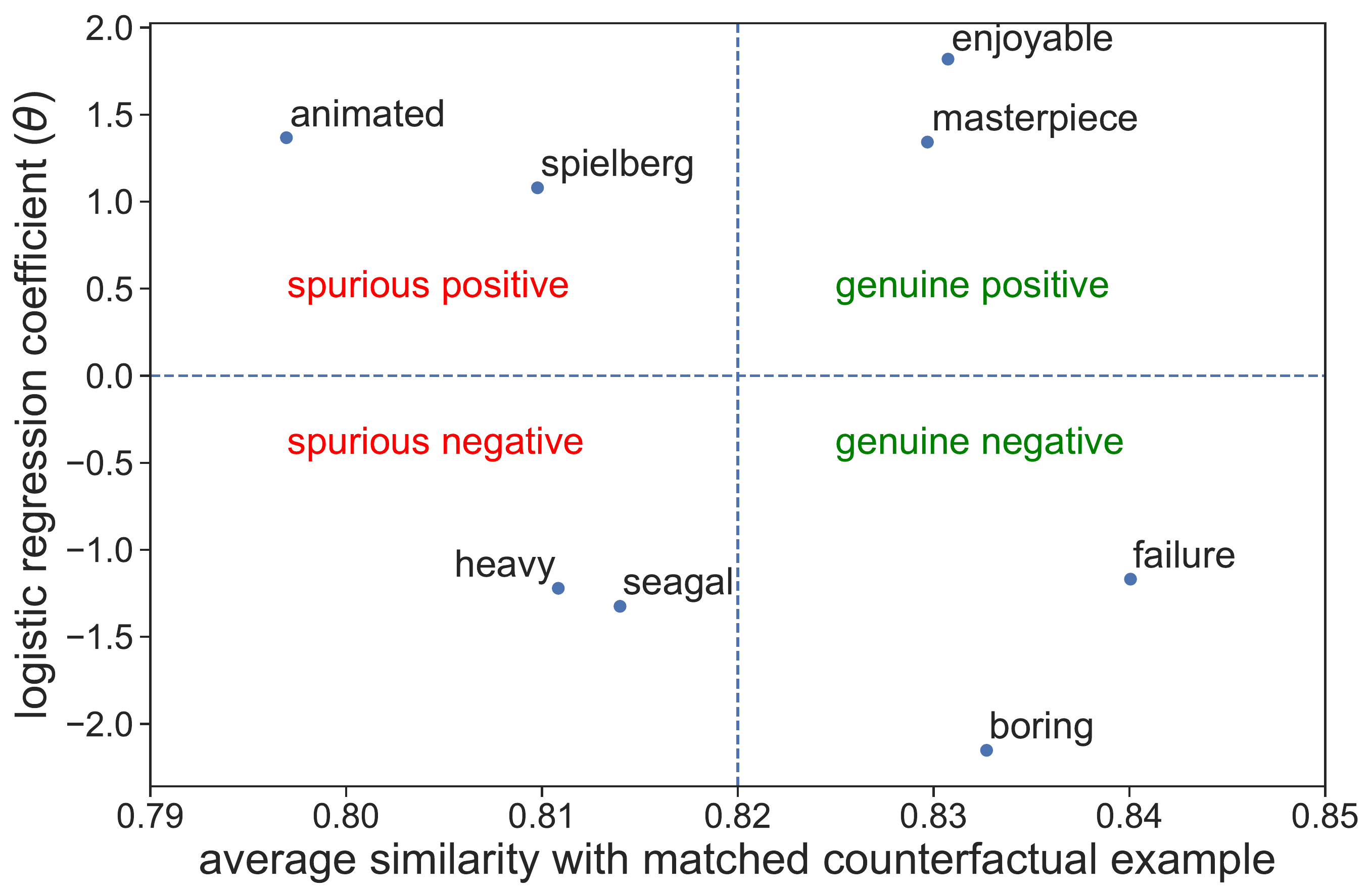}
        \caption{Motivating example of spurious and genuine correlations in a sentiment classification task.\label{fig.motivation}}
\end{figure}

In Figure~\ref{fig.motivation}, we show eight words with high magnitude coefficients for a classifier fit on a dataset of movie reviews~\cite{pang2005seeing}, where class $1$ means positive sentiment and $-1$ means negative sentiment. We will return shortly to the meaning of the $x$-axis; for now, let us consider the $y$-axis, which is the estimated coefficient $\theta_w$ for each word. Of the four words strongly correlated with the positive class ($\theta_w > 0$), two seem genuine ({\it enjoyable}, {\it masterpiece}), while two seem spurious ({\it animated}, {\it spielberg}). (Steven Spielberg is a very successful American director and producer.) Similarly, of the words correlated with the negative class, two seem genuine ({\it boring}, {\it failure}) and two seem spurious ({\it heavy}, {\it seagal}). (Steven Seagal is an American actor mostly known for martial-arts movies.) Furthermore, in some cases, the spurious term actually has a {\it larger} magnitude of coefficient than the genuine term (e.g., {\it seagal} versus {\it failure}).


Our goal in this paper is to distinguish between spurious and genuine correlations. Without wading into long-standing debates over the nature of causality~\cite{aldrich1995correlations}, 
we simplify the distinction between genuine and spurious correlations as a dichotomous decision: the discovered relationship between word $w$ and label $y$ is genuine if, all else being equal, one would expect $w$ to be a determining factor in assigning a label to a sentence. We use human annotators to make this distinction for training and evaluating models.

In this light, our problem is related to prior work on active learning with rationales~\cite{zaidan2007using,sharma2015active} and interactive feature selection~\cite{raghavan2005interactive}. However, our goal is not solely to improve prediction accuracy, but also to improve robustness across different groups affected by these spurious correlations. 


\section{Methods}
\label{sec.methods}

Our definition of {\it genuine} correlation given above fits well within the counterfactual framework of causal inference~\cite{winship1999estimation}. If the word $w$ in $s$ were replaced with some other word $w'$, how likely is it that the label $y$ would change? Since conducting randomized control trials to answer this counterfactual for many terms and sentences is infeasible, we instead resort to matching methods, commonly used to estimate average treatment effects from observational data~\cite{imbens2004nonparametric,king2019propensity}. The intuition is as follows: if $w$ is a reliable piece of evidence to determine the label of $s$, we should be able to find a very similar sentence $s'$ that (i) does not contain $w$, and (ii) has the opposite label of $s$. While this is not a {\it necessary} condition (there may not be a good match in a limited training set), in the experiments below we find this to be a fairly precise approach to identify genuine correlations.

\citenoun{paul2017feature} proposed a similar formulation, using propensity score matching to estimate the treatment effect for each term, then performing feature selection based on these estimates. Beyond recent critiques of propensity scores~\cite{king2019propensity}, any matching approach will create matches of varying quality, making it difficult to distinguish between spurious and genuine correlations. Returning to Figure~\ref{fig.motivation}, the $x$-axis shows the average quality of the counterfactual match for each term, where a larger value means that the linguistic context of the counterfactual sentence is very similar to the original sentence. (These are computed by cosine similarity of sentence embeddings, described in \S\ref{match}.) Even though these terms have very similar average treatment effect estimates, the quality of the match seems to be a viable signal of whether the term is spurious or genuine.

More generally, building on prior work that treats causal inference as a classification problem~\cite{lopez2015towards}, we can derive a number of features from the components of the treatment effect estimates (enumerated in \S\ref{sec.spurious_correlation}), and from these fit a classification model to determine whether a word should be labeled as spurious or genuine. This {\it word classifier} can then be used in a number of ways to improve the document classifier (e.g., to inform feature selection, to place priors on word coefficients, etc.).

To build the word classifier, we assume a human has annotated a small number of terms as spurious or genuine, which we can use as training data. While this places an additional cost on annotation, the nature of the features reduces this burden --- there are not very many features in the word classifier, and they are mostly generic / domain independent. As a result, in the experiments below, we find that useful word classifiers can be built from a small number of labeled terms (200-300). Furthermore, and perhaps more importantly, we find that the word classifier can be transported to new domains with little loss in accuracy. This suggests that one can label words once in one domain, fit a word classifier, and apply it in new domains without annotating additional words.

\subsection{Overview of approach}
The main stages of our approach are as follows:
\begin{packed_enum}
\item Given training data $\mathcal{D}=\{(s_1,y_1) \ldots (s_n,y_n)\}$ for the primary classification task, fit an initial classifier $f(x;\theta)$.
\item Extract from $f(x;\theta)$ the words $\mathcal{W}=\{w_1 \ldots w_m\}$ that are most strongly associated with each class according to the initial classifier. E.g., for logistic regression, we may extract the words with the highest magnitude coefficients for each class. For more complex models, other transparency algorithms may be used~\cite{martens2014explaining}.
\item For each word, compute features that indicate its likelihood to be spurious or genuine (\S\ref{sec.spurious_correlation}).
\item Fit a word classifier $h(w;\lambda)$ on a human-annotated subset of $\mathcal{W}$.
\item Apply $h(w;\lambda)$ on remaining words to estimate the probability that they are spurious.
\end{packed_enum}

After the final step, one may use the posterior probabilities in several ways to improve classification. E.g., to sort terms for feature selection, to place priors on word coefficients, to set attention weights in neural networks, etc. In this paper, we focus on feature selection, leaving other options for future work.

Additionally, we experiment with domain adaptation, where $h(w;\lambda)$ is fit on one domain and applied to another domain for feature selection, without requiring additional labeled words from that domain.

\subsection{Matching}
\label{match}

Most of the features for the word classifier are inspired by matching approaches from causal inference~\cite{stuart2010matching}. The idea is to match sentences containing different words in similar contexts so that we can isolate the effect that one particular word choice has on the class label.

For a word $w$ and a sentence $s$ containing this word, we let $s[{\hat w}]$ be the sentence $s$ with word $w$ removed. The goal of matching is to find some other context $s'[\hat w']$ such that $w \notin s'$ and $s[{\hat w}]$ is semantically similar to $s'[{\hat w'}]$. We use a {\it best match} approach, finding the closest match $s^* \leftarrow \argmax_{s'} \hbox{sim}(s[{\hat w}], s'[{\hat w'}])$. With this best match, we can compute the average treatment effect (ATE) of word $w$ in $N$ sentences:
\begin{equation}
\label{eq.ate}
\tau_w = \frac{1}{N}\sum_{\{s | w \in s\}} y_s - y_{s^*}
\end{equation}
Thus, a term $w$ will have a large value of $\tau_w$ if (i) it often appears in the positive class, and (ii) very similar sentences where $w$ is swapped with $w'$ have negative labels.

In our experiments, to improve the quality of matches, we limit contexts to the five previous and five subsequent words to $w$, then represent the context by concatenating the last four layers of a pre-trained BERT model (recommended by the original BERT paper)~\cite{Devlin2018BERT}. We use the cosine similarity of context embeddings as a measure of semantic similarity. 


Take one example from Table \ref{tab.matches}: {\it ``it's \underline{refreshing} to see a movie that (1)"} is matched with {\it ``it's \underline{rare} to see a movie that (-1)"}. Words {\it refreshing} and {\it rare} appear in similar contexts, but adding {\it refreshing} to this context makes the sentence positive, while adding {\it rare} to this context makes it negative. If most of the pairwise matches show that adding {\it refreshing} is more positive than adding other substitution words, then {\it refreshing} is very likely to be a genuine positive word.

On the contrary, if adding other substitution words for similar contexts does not change the label, then $w$ is likely to be a spuriously correlated word. Take another example from Table \ref{tab.matches}, {\it ``smoothly under the direction of \underline{spielberg} (1)"} is matched with {\it ``it works under the direction of \underline{kevin} (1)"}, {\it spielberg} and {\it kevin} appear in similar contexts, and substituting {\it spielberg} with {\it kevin} does not make any difference in the label. If most pairwise matches show that substituting {\it spielberg} to other words does not change the label, then {\it spielberg} is very likely to be a spurious positive word.


\subsection{Features for Word Classification}
\label{sec.spurious_correlation}

\begin{table}[t]
    \small
    \centering
    \begin{tabular}{ | l |} 
     \hline
    it's {\bf refreshing} to see a movie that (1) \\ it's {\bf rare} to see a movie that (-1) \\
    \hline
    cast has a lot of {\bf fun} with the material (1) \\ comedy with a lot of {\bf unfunny} (-1) \\
    \hline
    smoothly under the direction of {\bf spielberg} (1) \\ it works under the direction of {\bf kevin} (1) \\
    \hline
    refreshingly different slice of asian {\bf cinema} (1)	\\  an interesting slice of {\bf history} (1) \\
    \hline
    charting the rise of hip-hop {\bf culture} in general (1) \\ hip-hop has a {\bf history}, and it's a metaphor (1) \\
    \hline
    \end{tabular}
    \caption{Examples of matched contexts from IMDB dataset; word substitutions are shown in bold.}
    \label{tab.matches}
\end{table}

While the matching approach above is a traditional way to estimate the causal effect of a word $w$ given observational data, there are many well-known limitations to matching approaches~\cite{Gary2011Matching}. A primary difficulty is that high-quality matches may not exist in the data, leading to biased estimates. Inspired by supervised learning approaches to causal inference~\cite{lopez2015towards}, rather than directly use the ATE to distinguish between spurious and genuine correlations, we instead compute a number of features to summarize information about the matching process.  In addition to the ATE itself, we calculate the following features:
\begin{packed_item}
    \item The average context similarity of every match for word $w$.
    \item The context similarity of the top-5 closest matches.
    \item The maximum and standard deviation of the similarity score.
    \item The context similarity of the closest positive and negative sentences.
    \item The weighted average treatment effect, where Eq. (\ref{eq.ate}) is weighted by the similarity between $s$ and $s^*$.
    \item The ATE restricted to the top-5 most similar matches for sentences containing $w$.
    \item The word's coefficient from the initial sentence classifier.
    \item Finally, to capture subtle semantic differences between the original and matched sentences, we compute features such as the average embedding difference from all matches, the top-3 most different dimensions from the average embedding, and the maximum value along each dimension. 
\end{packed_item}

\subsection{Measuring the Impact of Spurious Correlations on Classification}
\label{sample_groups}

After we train the word classifier to identify spurious and genuine words, we are further interested in exploring how spurious correlations affect classification performance on test data. As discussed in \S\ref{sec.intro}, measuring robustness can be difficult when data are sampled i.i.d. because the same spurious correlations exist in the training and testing data. Thus, we would not expect accuracy to necessarily improve on a random sample when spurious words are removed. Instead, we are interested in measuring the {\it robustness} of the classifier, where robustness is with respect to which subgroup of data is being considered.



Motivated by~\cite{sagawa2020distributionally}, we divide the test data into two groups and explore the model performance on each. The first group, called the {\it minority group}, contains sentences in which the spurious correlation is expected to mislead the classifier. From our running example, that would be a {\it negative} sentiment sentence containing {\it spielberg}, or a {\it positive} sentiment sentence containing {\it seagal}. Analogously, the {\it majority group} contains examples in which the spurious correlation helps the classifier (e.g., positive sentiment documents containing {\it spielberg}). In \S\ref{feature_selection_strategies}, we conduct experiments to see how removing terms that are predicted to be spurious could affect accuracies on majority and minority groups.

\section{Experiments}
\label{sec.experiments}


\subsection{Data}
\label{sec.data}

We experiment with four datasets for two binary classification tasks: sentiment classification and toxicity detection.\footnote{Code and data available at: \url{https://github.com/tapilab/emnlp-2020-spurious}}

\begin{packed_item}
    \item {\it IMDB movie reviews}: movie review sentences labeled with their overall sentiment polarity (positive or negative)~\cite{pang2005seeing} (version 1.0).
    \item {\it Kindle reviews}: product reviews from Amazon Kindle Store with ratings range from 1-5~\cite{He_2016}. We first fit a sentiment classifier on this dataset to identify keywords, and then split each review into single sentences and assign each sentence the same rating as the original review. We select sentences that contain sentiment keywords and then remove sentences that have fewer than 5 or more than 40 words, and finally label the remaining sentences rated \{4,5\} as positive and sentences rated \{1,2\} as negative. To justify the validity of sentence labels inherited from original documents, we randomly sampled 500 sentences (containing keywords) and manually checked their labels. The inherited labels were correct for 484 sentences (i.e., 96.8\% accuracy).
    
    \item {\it Toxic comment}: a dataset of comments from Wikipedia's talk page~\cite{Wulczyn2017Toxic}.\footnote{\kaggleurl} Comments are labeled by human raters for toxic behavior (e.g., comments that are rude, disrespectful, offensive, or otherwise likely to make someone leave a discussion). Each comment was shown up to 10 annotators and the fraction of human raters who believed the comment is toxic serves as the final toxic score that ranges from 0.0 to 1.0. We follow the same processing steps in {\it Kindle reviews dataset}: split comments into sentences, select sentences containing toxic keywords (learned from a toxic classifier), and limit sentence length. We label sentences with toxicity scores $\ge 0.7$ as toxic and $\le 0.5$ as non-toxic. 
    
    \item {\it Toxic tweet}: tweets collected through Twitter Streaming API by matching toxic keywords from HateBase and labeled as toxic or non-toxic by human raters~\cite{Bahar2020ICWSM}. 
\end{packed_item}

    

All datasets are sampled to have an equal class balance. The basic dataset information is summarized in Table \ref{tab.corpus}.

\begin{table}[t]
    \small
    \centering
    \begin{tabular}{ | r | c | c | c |} 
     \hline
     & \makecell{\#docs \\ \\} & \makecell{\#top \\ words} & \makecell{\#matched \\ sentences} \\
     \hline
    IMDB & 10,662 & 366 & 8,882 \\
    Kindle & 20,232 & 270 & 24,882 \\
    Toxic comment & 15,216 & 329 & 8,414 \\
    Toxic tweet & 6,774 & 341 & 9,224 \\
    \hline
    \end{tabular}
    \caption{Corpus summary}
    \label{tab.corpus}
\end{table}

\subsection{Creating Matched Sentences}
We first get pairwise matched sentences for words of interest. In this work, we focus on words that have relatively strong correlations with each class. So we fit a logistic regression classifier for each dataset and select the top features by placing a threshold on coefficient magnitude (i.e., words with high positive or negative coefficients). For IMDB movie reviews, Kindle reviews, and Toxic comments, we use a coefficient threshold 1.0; and for Toxic tweet, we use threshold 0.7 (to generate a comparable number of candidate words).


We find matched sentences for each word following the method in \S\ref{match}. Table~\ref{tab.matches} shows five examples of pairwise matches. The total number of matched sentences are shown in Table~\ref{tab.corpus}. 




\subsection{Word Classification}

The goal of word classification is to distinguish between spurious words and genuine words.  We first manually label a small set of words as spurious or genuine (Table \ref{tab.word_clf_evaluation}). For sentiment classification, we consider both positive and negative words. For toxicity classification, we only consider toxic words. We had two annotators annotate each term; the agreement was generally high for this task (e.g., 96\% raw agreement), with the main discrepancies arising from the knowledge of slang and abbreviations. 

We represent each word with the numerical features calculated from matched sentences (\S\ref{sec.spurious_correlation}), standardized to have zero mean and unit variance. Finally, we apply a logistic regression model for the binary word classifier\cut{~\cite{scikit-learn}}. We explore the word classifier performance for the same domain and domain adaptation. 

{\bf Same domain:} We apply 10-fold cross-validation to estimate the word classifier's accuracy within the same domain. In practice, the idea is that one would label a set of words, fit a classifier, then apply to the remaining words.

{\bf Domain adaptation:} To reduce the word annotation burden, we are interested in understanding whether a word classifier trained on one domain can be applied in another. Thus, we measure cross-domain accuracy, e.g., by fitting the word classifier on IMDB dataset and evaluating on Kindle dataset.


\subsection{Feature Selection Based on Spurious Correlation}
\label{feature_selection_strategies}
We compare several strategies to do feature selection for the initial document classification tasks.

According to the word classifier, each word is assigned a probability of being spurious, which we use to sort terms for feature selection. That is, words deemed most likely to be spurious are removed first. As a comparison, we experiment with the following strategies to rank words in the order of being removed. 

{\bf Oracle} This is the gold standard. We treat the manually labeled spurious words as equally important and sort them in random order. This gold standard ensures that the removed features are definitely spurious.

{\bf Sentiment lexicon} We create a sentiment lexicon by combing sentiment words from~\cite{Wilson2005Sentiment} and ~\cite{Liu2012Sentiment}. It contains 2724 positive words and 5078 negative words. We select words that appear in the sentiment lexicon as informative genuine features and fit a baseline classifier with these features. This is a complementary method with the previous method by oracle.

{\bf Random} This is a baseline method that sorts the top words in random order, where top words could be spurious or genuine, and the words are removed in random order.

{\bf Same domain prediction} We sort words in descending order of the probability of being spurious, according to the word classifier trained on the same domain (using cross-validation).


{\bf Domain adaptation prediction} This is a similar sorting process with the previous strategy except that the probability is from domain adaptation, where the word classifier is trained on a different dataset. We consider domain transfer between IMDB and Kindle datasets, and between Toxic comment and Toxic tweet datasets.

In the document classification task, we sample majority and minority groups by selecting an equal number of sentences for each top word to ensure a fair comparison during feature selection. We check feature selection performance for each group by gradually removing spurious words following the order of each strategy described above. As a final comparison, we also implement the method suggested in~\citenoun{sagawa2020investigation}, which reduces the effect of spurious correlation from training data. To do so, we sample the majority and minority group from training data, and down-sample the majority group to have an equal size with the minority group. We then fit the document classifier on the new training data and evaluate its performance on the test set. Note that this method assumes knowledge of which features are spurious. Our approach can be seen as a way to first estimate which features are spurious and then adjust the classifier accordingly.





\section{Results and Discussion}
\label{sec.results}
In this section, we show results for identifying spurious correlations and then analyze the effect of removing spurious correlations in different cases.

\subsection{Word Classification}
\label{word_clf_results}
Table \ref{tab.word_clf_evaluation} shows the ROC AUC scores for classifier performance. To place these numbers in context, recall that the words being classified were specifically selected because of their strong correlation with the class labels. For example, some spurious positive words appear in 20 positive documents and only a few negative documents. Despite the challenging nature of this task, Table \ref{tab.word_clf_evaluation} shows that word classifier performs well at classifying spurious and genuine words with AUC scores range from 0.657 to 0.823. Furthermore, the domain adaptation results indicate limited degradation in accuracy, and occasionally improvements in accuracy. The exception is the toxic tweet dataset, where the score is 6\% worse for domain adaptation. We suspect that this is caused by the low-quality texts in the toxic tweet dataset (this is the only dataset that the text is tweets instead of formal sentences). 

\begin{figure}[t]
	\centering
	\includegraphics[width=3.0in]{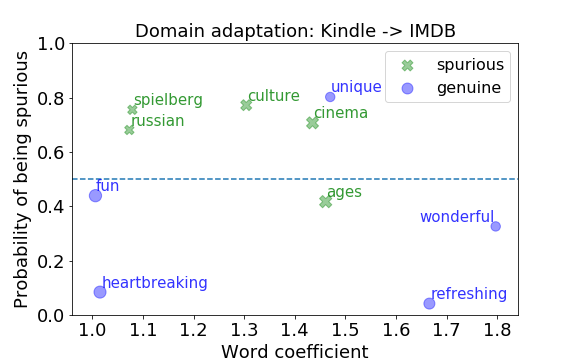}
	\caption{Example of spurious and genuine words predicted by the word classifier trained on words from Kindle reviews and applied to words from IMDB reviews.
	}
	\label{fig:wd_clf}
\end{figure}

 
Fig \ref{fig:wd_clf} shows an example of the domain adaptation results. We observe that {\it culture, spielberg, russian, cinema} are correctly predicted to have high probabilities of being spurious, while {\it refreshing, heartbreaking, wonderful, fun} are correctly predicted to have relatively lower probabilities of being spurious. We also observe that the predictions for {\it unique} and {\it ages} do not agree with human labels. We show top-5 spurious and genuine words predicted for each dataset in Table~\ref{tab.word_clf_topTerms}. Error analysis suggests that misclassifications are often due to small sample sizes -- some genuine words simply do not appear enough to find good matches. In future work, we will investigate how data size influences accuracy.

\begin{table}[t]
    \small
    \centering
    \begin{tabular}{ | c | c | c || c | c |} 
     \hline
     & \makecell{\bf IMDB \\ reviews} & \makecell{\bf Kindle \\ reviews} & \makecell{\bf Toxic \\  comment} & \makecell{\bf Toxic \\ tweet}\\
     \hline
    \#spurious & 90 & 119 & 40 & 72 \\
    \#genuine & 174 & 100 & 73 & 45 \\
    \hline
    \hline 

    \makecell{same \\ domain} & 0.776 & 0.657 & 0.823 & 0.686 \\
    \hline
    \makecell{ domain \\ adaptation} & 0.741 & 0.699 & 0.726 & 0.744 \\
    \hline
    
    \end{tabular}
    \caption{Word classifier performance (AUC score)}
    \label{tab.word_clf_evaluation}
\end{table}

Examining the top coefficients in the word classifier, we find that features related to the match quality tend to be highly correlated with genuine words (e.g., the context similarity of close matches, ATE calculated from the close matches). In contrast, features calculated from the embedding differences of close matches have relatively smaller coefficients.\footnote{Detailed feature coefficients and analysis of feature importance are available in the code.} For example, in the word classifier trained for IMDB dataset, the average match similarity score has a coefficient of 1.3, and the ATE feature has a coefficient of 0.8. These results suggest that the quality of close matches is viable evidence of genuine features, and combining traditional ATE estimates with features derived from the matching procedure can provide stronger signals for distinguishing spurious and genuine correlations.


\begin{table*}[t]
    \small
    \centering
    \begin{tabular}{ |p{1.8cm}|p{1.7cm}||p{1.6cm}|p{1.7cm}||p{1.8cm}|p{1.3cm}||p{1.4cm}|p{1.2cm}|} 
     \hline
       \multicolumn{2}{| c ||}{\textbf{IMDB}} & \multicolumn{2}{ c ||}{\textbf{Kindle}} & \multicolumn{2}{ c ||}{\textbf{Toxic comment}}  & \multicolumn{2}{ c |}{\textbf{Toxic tweet}} \\
     \hline
       \textbf{spurious} & \textbf{genuine} & \textbf{spurious} & \textbf{genuine} & \textbf{spurious} & \textbf{genuine} & \textbf{spurious} & \textbf{genuine}\\
     \hline
     unintentional & refreshing & boy & omg & intelligence & idiot & edkrassen & cunt \\ 
    
     russian & horrible & issues & definitely & \st{parasites} & stupid & hi & twat \\ 
    
    benigni & uninspired & \st{benefits} & \st{draw} & \st{sucking} & idiots & \st{pathetic} & retard \\ 
   
    animated & strength & \st{teaches} & returned & mongering & stupidity & side & pussy \\ 
    
    pulls & exhilarating & girl & \st{halfway} & lifetime & moron & example & ass \\ 
    \hline\hline
    \st{visceral} & refreshing & finds & omg & mongering & stupid & aint & cunt \\ 
    
    mike & rare & mother & highly & \st{lunatics} & idiot & between & twat \\ 
    
    unintentional & horrible & girl & returned & \st{slaughter} & idiots & wet & retard \\ 
    
    \st{strange} & ingenious & us & \st{down} & narrative & idiotic & side & faggot \\ 
    
    \st{intelligent} & sly & humans & enjoyed & brothers & stupidity & rather & pussy \\ 
    \hline
    \end{tabular}
    
    \caption{Top 5 spurious and genuine words predicted by the in-domain word classifier (first five rows) and cross-domain classifier (last five rows). Words with strike-through are incorrectly classified.
    }
    \label{tab.word_clf_topTerms}
\end{table*}

    
  
    

\subsection{Feature Selection by Removing Spurious Correlations}
\label{feature_selection_result}

\begin{figure}[t]
	\centering
	\includegraphics[width=3.0in]{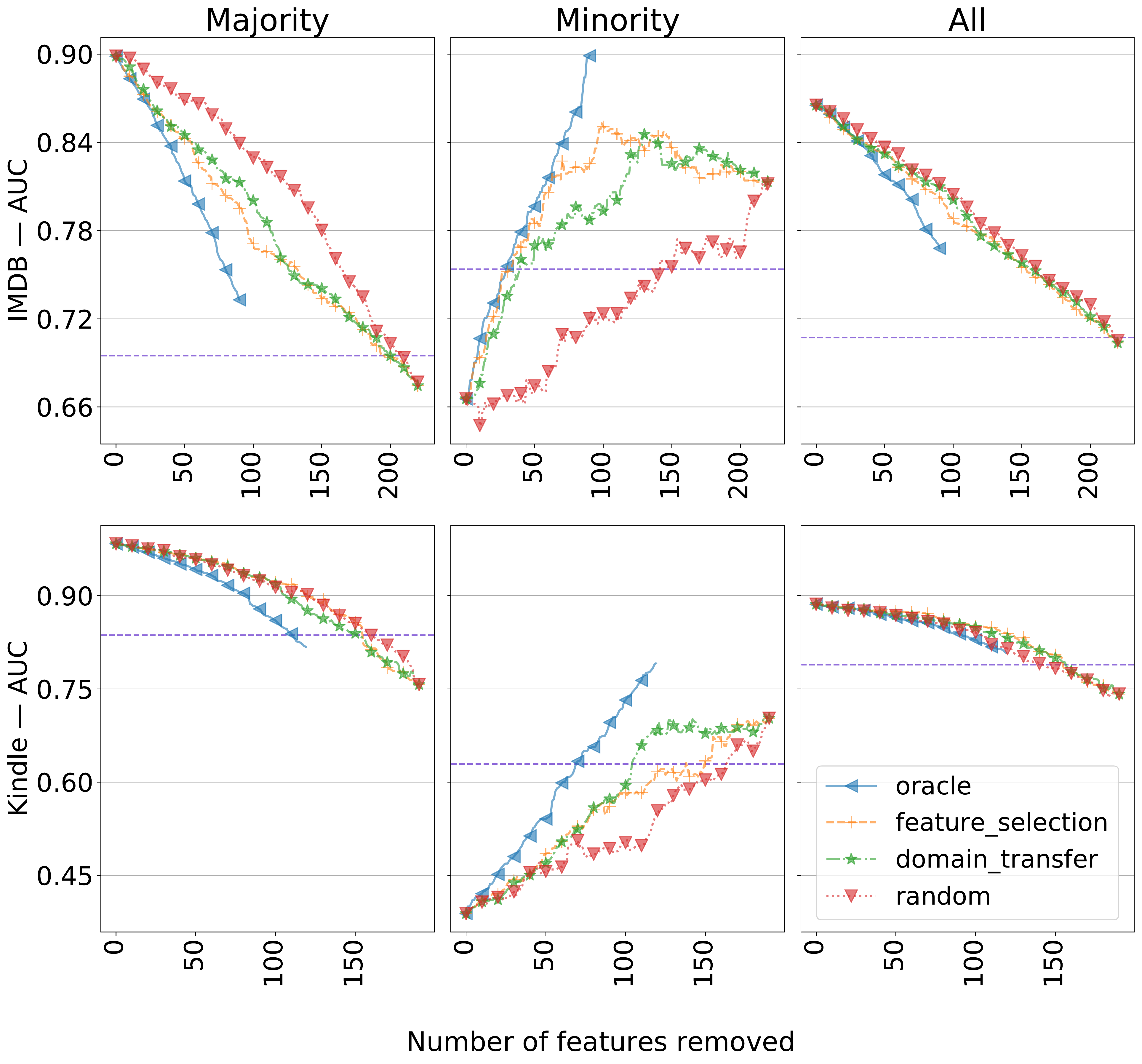}
	\caption{Feature selection for sentiment classification.}
	\label{fig:feature_selection_sentiment}
\end{figure}

We apply different feature selection strategies in \S\ref{feature_selection_strategies} and test the performance on majority set, minority set, and the union of majority and minority sets (denoted as ``All''). 

Fig~\ref{fig:feature_selection_sentiment} shows feature selection performance on IMDB movie reviews and Kindle reviews. The starting point in each plot shows the performance of not removing any feature. The horizontal line in-between shows the performance of the method suggested in~\citenoun{sagawa2020investigation}. 

For the majority group, because spurious correlations learned during model training agree with sentence labels, the model performs well on this group, and removing spurious features hurts performance (i.e., about 20\% drop of AUC score in both datasets). On the contrary, the spurious correlations do not hold in the minority group. Thus, the model does not perform well at the starting point when not removing any spurious feature, and the performance increases when we gradually remove spurious features. After removing enough spurious features, the model performance stabilizes.

For IMDB reviews, removing spurious features improves performance by up to 20\% AUC for the minority group, and feature selection based on predictions from the word classifier outperforms random ordering substantially. For Kindle, removing spurious features improves accuracy by up to 30\% AUC for the minority group. Interestingly, domain adaptation actually appears to outperform the within-domain results, which is in line with word classifier performance shown in Table~\ref{tab.word_clf_evaluation} (i.e., domain adaptation outperforms within domain AUC by 4.2\% for Kindle word classifier). The result on ``All" shows the trade-off between the performance on the majority group and minority group. If removing spurious features hurts more on the majority group than it helps the minority group, then the performance on the ``All'' set would decrease, and vice versa. In our experiment, the majority group has more samples than the minority group, so the final performance on the ``All'' set gradually decreases when removing spurious features. 

\begin{figure}[t]
	\centering
	\includegraphics[width=3.02in]{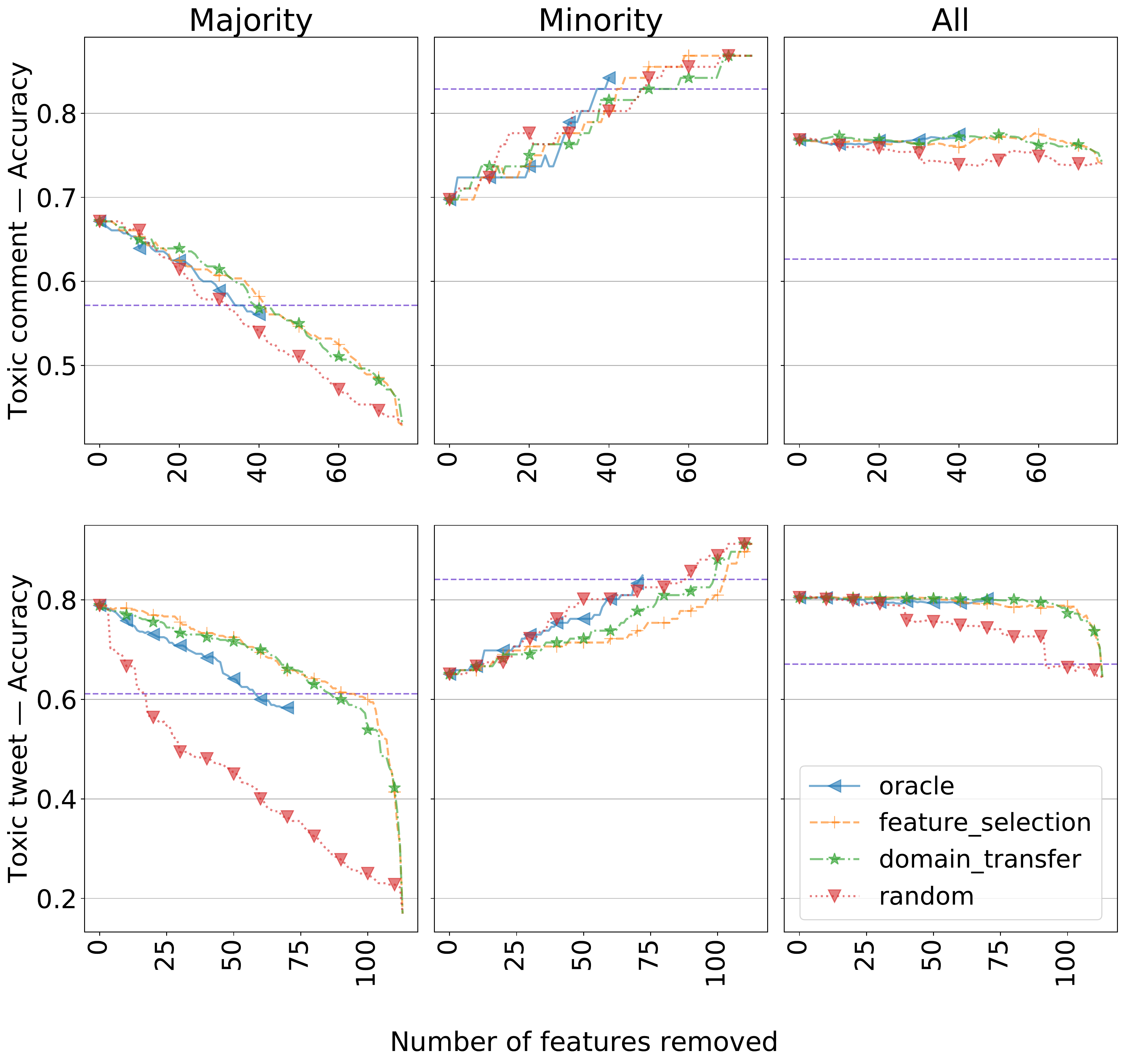}
	\caption{Feature selection for toxicity classification. Test sets are selected with respect to toxic features, so there's only one class for each set. We show accuracy score on y-axis.}
	\label{fig:feature_selection_toxic}
\end{figure}

We also perform feature selection on Toxic comment and Toxic tweet datasets, where we only focus on toxic features. As shown in Fig~\ref{fig:feature_selection_toxic}, for the minority set, removing spurious features improves performance by up to 20\% accuracy for Toxic comment, and 30\% accuracy for Toxic tweet. Compared with sentiment datasets, toxic datasets have fewer spurious words to remove because we only cares about spurious toxic features and don't care about non-toxic features. While in sentiment classification, the spurious words are from both positive and negative classes. Besides that, the Toxic tweet dataset is noisy with low-quality texts. So the feature selection methods perform differently on toxic datasets compared with sentiment datasets.


Additionally, the baseline method of using sentiment lexicon has limited contribution (e.g., performance scores for different datasets are: IMDB, 0.776; Kindle, 0.636; Toxic comment, 0.592; Toxic tweet, 0.881;), which is about 0.05 to 0.2 lower compared with the performance of the proposed feature selection methods. The reasons are: (i) the sentiment lexicon missed some genuine words that are specific to each dataset (e.g., {\it `typo'} is a negative word when used in kindle book reviews but is missed from the sentiment lexicon); (ii) the same word might convey different sentiments depending on the context. E.g., {\it `joke'} is positive in {\it ``He is humorous and always tell funny jokes''}, but is negative in {\it ``This movie is a joke''}; (iii) in the toxic classification task, there's no direct relation between toxicity and sentiment. A toxic word can be positive and a non-toxic word can be negative (e.g., {\it `unhappy'}). Instead of using sentiment lexicon, this paper aims to create a method to automatically identify genuine features that are specific to each dataset, and this method could generalize to different tasks in addition to sentiment classification.





\section{Related Work}
\label{sec.related}

\citenoun{wood2018challenges} and \citenoun{keith2020text} provide good overviews of the growing line of research combining causal inference and text classification. 
Two of the most closely related works mentioned previously, are  \citenoun{sagawa2020investigation} and \citenoun{paul2017feature}.

\citenoun{sagawa2020investigation} investigates how spurious correlations arise in classifiers due to overparameterization. They compare overparameterized models with underparameterized models and show that overparameterization hurts worst-group error, where the spurious correlation does not hold. They do simulation experiments with core features encoding actual label and spurious features encoding spurious attributes. Results show that the relative size of the majority group and minority group as well as the informativeness of spurious features modulate the effect of overparameterization. While \citenoun{sagawa2020investigation} assumes it is known ahead of time which features are spurious, here we instead try to predict that in a supervised learning setting.

\citenoun{paul2017feature} proposes to do feature selection for text classification 
by causal inference. He adapts the idea of propensity score matching to document classification and identifies causal features from matched samples. Results show meaningful word features and interpretable causal associations. Our primary contributions beyond this prior work are (i) to use features of the matching process to better identify spurious terms using supervised learning, and (ii) to analyze effects in terms of majority and minority groups. Indeed, we find that using the treatment effect estimates alone for the word classifier results in worse accuracy than combining it with the additional features.



Recently, \citenoun{kaushik2020learning} show the prevalence of spurious correlations in machine learning by having humans make minimal edits to change the class label of a document. Doing so reveals large drops in accuracy due to the model's overdependence on spurious correlations.

Another line of work investigates how confounds can lead to spurious correlations in text classification~\cite{elazar2018adversarial,Landeiro_2018,pryzant2018deconfounded,garg2019counterfactual}. These methods typically require the confounding variables to be identified beforehand (though \citenoun{kumar2019topics} is an exception).

A final line of work views spurious correlations as a result of an adversarial, data poisoning attack~\cite{chen2017targeted,Dai+Chen:2019}. The idea is that an attacker injects spurious correlations into the training data, so as to control the model's predictions on new data. While most of this research focuses on the nature of the attack models, future work may be able to combine the approaches in this paper to defend against such attacks.


\section{Conclusion}
\label{sec.conclusions}

We have proposed a supervised classification method to distinguish spurious and genuine correlations in text classification. Using features derived from matched samples, we find that this word classifier achieves moderate to high accuracy even tested on strongly correlated terms. Additionally, due to the generic nature of the features, we find that this classifier does not suffer much degradation in accuracy when trained on one dataset and applied to another dataset. Finally, we use this word classifier to inform feature selection for document classification tasks. Results show that removing words in the order of their predicted probability of being spurious results in more robust classification with respect to worst-case accuracy.


\section*{Acknowledgments}

This research was funded in part by the National Science Foundation under grants \#IIS-1526674 and \#IIS-1618244.

\bibliography{emnlp2020}
\bibliographystyle{acl_natbib}


\end{document}